\documentclass[runningheads]{llncs}
\pdfoutput=1

\usepackage{tkz-graph}
\usetikzlibrary{arrows}
\usepackage{tikz}
\usepackage{subfig}
\usetikzlibrary{arrows,automata}
\usepackage{enumerate}
\usepackage{scalefnt}

\def\clasp{{\sc Clasp}}
\def\lar{\leftarrow}
\def\beq{\begin{equation}}
\def\eeq#1{\label{#1}\end{equation}}
\def\ba{\begin{array}}
\def\ea{\end{array}}
\def\no{\ii{not}}
\def\ii#1{\hbox{\it #1\/}}
\def\is#1{\hbox{\scriptsize\it #1\/}}
\def\aspI{{\ensuremath{\ii{ASP}_1}}}
\def\aspII{{\ensuremath{\ii{ASP}_2}}}
\def\aspIopt{{\ensuremath{\ii{ASP}_1^{\is{opt}}}}}
\def\aspIIopt{{\ensuremath{\ii{ASP}_2^{\is{opt}}}}}
\def\satI{{\ensuremath{\ii{SAT}_1}}}
\def\satII{{\ensuremath{\ii{SAT}_2}}}
\def\bforce{{\ensuremath{\ii{BF}}}}

\begin{document}

\title{Generating Shortest Synchronizing Sequences using Answer Set Programming}
\titlerunning{Generating Shortest Synchronizing Sequences using ASP}

\author{Canan G\"uni\c{c}en \and Esra Erdem \and H\"usn\"u Yenig\"un}
\authorrunning{C.~G\"uni\c{c}en \emph{et al}\/.} %C. G\"uni\c{c}en et al.} % abbreviated author list (for running head)
\setcounter{page}{117}

\institute{Sabanci University, Orhanli Tuzla, Istanbul 34956, Turkey\\
\email{\{canangunicen,esraerdem,yenigun\}@sabanciuniv.edu}}

\maketitle

% % % % % % % % % % % % % % % % % % % % % % % % % % %

\begin{abstract}
For a finite state automaton, a synchronizing sequence is an input
sequence that takes all the states to the same state. Checking the
existence of a synchronizing sequence and finding a synchronizing
sequence, if one exists, can be performed in polynomial time.
However, the problem of finding a shortest synchronizing sequence is
known to be NP-hard. In this work, the usefulness of Answer Set
Programming to solve this optimization problem is investigated, in
comparison with brute-force algorithms and SAT-based approaches.

\end{abstract}

% % % % % % % % % % % % % % % % % % % % % % % % % % %

\section{Introduction}

For a state based system that reacts to events from its environment
by changing its state, a synchronizing sequence is a specific
sequence of events that brings the system to a particular state no
matter where the system initially is. Synchronizing sequences have
found applications in many practical settings. In model based
testing, it can be used to bring the unknown initial state of an
implementation to a specific state to start
testing~\cite{LY94,KJ10}. As Natarajan~\cite{Natarajan86} and
Eppstein~\cite{Eppstein90} explain, it can be used to orient a part
to a certain position on a conveyor belt. Another interesting
application is from biocomputing, where one can use a DNA molecule
encoding a synchronizing sequence to bring huge number of identical
automata (in the order of $10^{12}$ automata/$\mu l$) to a certain
restart state~\cite{AV04}.

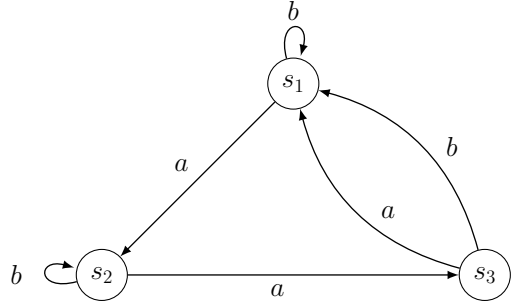
\begin{figure}
%\begin{center}
\centering
\begin{tikzpicture}[->,node distance=6cm, scale=0.6,transform shape]
  \tikzstyle{every state}=[draw,shape=circle, text=black,minimum size=30pt]
    \scalefont{2}
  \tikzstyle{selected edge} = [->,>=latex,draw,line width=0.5pt,black]
  \tikzstyle{ignored edge} =  [->,>=latex,draw,line width=0.7pt,black!35]
  \node[state]         (A)                    {$s_1$};
  \node[state]         (B) [below left of=A] {$s_2$};
  \node[state]         (C) [below right of=A] {$s_3$};

 \path [selected edge](A)
            edge [loop above] node {$b$}  (A)
            edge [] node [above,xshift=-10pt] {$a$} (B)

 (B)        edge [] node [below,xshift=-10pt] {$a$} (C)
            edge [loop left]  node [left,xshift=-10pt] {$b$} (B)
 (C)
            edge [bend right] node [right,xshift=10pt] {$b$} (A)
            edge [bend left]  node [right,xshift=10pt] {$a$} (A);
\end{tikzpicture}
\caption{An example FA $A_1$} \label{fig:A1}
%\end{center}
\end{figure}

Such a state based system can be formalized as a {\em finite state
automaton}~(FA). We restrict ourselves to {\em deterministic} FA,
which is defined as a tuple $A = (Q,\Sigma,\delta)$, where $Q$ is a
finite set of states, $\Sigma$ is a finite input alphabet, and
$\delta: Q \times \Sigma \mapsto Q$ is a transition function,
defining how each state of $A$ is changed by the application of
inputs. The transition function $\delta$ is extended to words in
$\Sigma^\star$ naturally as $\delta(q,\varepsilon)=q$, $\delta(q,w
x) = \delta(\delta(q,w),x)$, where $q \in Q$, $w \in \Sigma^\star$,
$x \in \Sigma$, and $\varepsilon$ is the empty word. A FA $A =
(Q,\Sigma,\delta)$ is called {\em completely specified} when
$\delta$ is a total function. We will only consider completely
specified FA in this work. Figure~\ref{fig:A1} is an example of a
FA.

We can now define a synchronizing sequence formally. Given an FA $A
= (Q,\Sigma,\delta)$, an input sequence $w \in \Sigma^\star$ is
called a {\em synchronizing sequence} for $A$ if $\forall q,q' \in
Q$, $\delta(q,w) = \delta(q',w)$. As an example, $baab$ is a
synchronizing sequence for $A_1$ given in Figure~\ref{fig:A1}.

Synchronizing sequences attracted much attention from a theoretical
point of view as well. In the literature, a synchronizing sequence
is also referred to as a synchronizing word, reset sequence, or a
reset word. Not every FA has a synchronizing sequence, and one can
check the existence of a synchronizing sequence for a given FA in
polynomial time. On the other hand, the problem of finding a
shortest synchronizing sequence is known to be
NP-hard~\cite{Eppstein90}. For this reason, several heuristic
approaches have been suggested to compute short synchronizing
sequences~\cite{Eppstein90,Trahtman04,Roman09,Kudlacik12}. These
algorithms guarantee a synchronizing sequence of length $O(n^3)$
where $n$ is the number of states in the FA. The best known upper bound is
$n(7n^2+6n-16)/48$~\cite{Trahtman11}. However, it has been
conjectured by {\v{C}}ern{\'y} almost half a century ago that this
upper bound is $(n-1)^2$~\cite{Cerny64,Cerny71} after providing a
class of FA with $n$ states whose shortest synchronizing sequence is
of length $(n-1)^2$. The conjecture is shown to hold for certain
classes of
automata~\cite{Eppstein90,AV04,Kari03,AV05,Trahtman07,Volkov09}.
However, the conjecture is still open in general, and it is one of
the oldest open problems of finite state automata theory.

Despite the fact that it is NP-hard, considering 
the computation of shortest synchronizing sequences is still useful. 
Such attempts are valuable both for understanding the characteristics of
shortest synchronizing sequence problem (see e.g.~\cite{Skv}) and
for forming a base line for the performance evaluation of heuristics for
computing short synchronizing sequences.

In this work, we formulate the problem of computing a shortest
synchronizing sequence in Answer Set Programming
(ASP)~\cite{lifschitz08,BrewkaET11}---a knowledge representation and
reasoning paradigm with an expressive formalism and efficient
solvers. The idea of ASP is to formalize a given problem as a
``program'' and to solve the problem by computing models (called
``answer sets''~\cite{gelfondL91}) of the program using ``ASP
solvers'', such as \clasp~\cite{gebserKNS07}.

%More precisely, given an FA $A = (Q,\Sigma,\delta)$, and a
%constant $c$, we generate an ASP formula which is satisfiable
%if and only if $A$ has a synchronizing sequence of length $c$.
%We also extend the SAT formulation of shortest synchronizing
%sequence problem presented in~\cite{Skv} to FA with more than two input
%symbols. We experimentally compare the performance of ASP
%solvers, SAT solvers and the brute--force algorithm to compute the
%shortest synchronizing sequence.

After we represent the shortest synchronizing sequence problem in
ASP, we experimentally evaluate the performance and effectiveness
of ASP, in comparison with two other approaches, one based on
SAT~\cite{Skv} and the other on a brute-force
algorithm~\cite{Chmiel:2010:CCP:1964285.1964294}. For our
experiments with the SAT-based approach, we extend the SAT
formulation of the existence of a synchronizing sequence of a given
length~\cite{Skv}, to FA with more than two input symbols.

The rest of the paper is organized as follows. In Section~\ref{sec:ASP-formulation}
we present four different ASP formulations for the problem.
An existing SAT formulation~\cite{Skv} is extended to FAs with more
than two inputs in Section~\ref{sec:SAT}. The experimental results
are given in Section~\ref{sec:experiments} to compare the approaches.
Finally, in Section~\ref{sec:conclusion} we give concluding remarks and
some future research directions.

% % % % % % % % % % % % % % % % % % % % % % % % % % %

\section{ASP Formulations of the Shortest Synchronizing Sequence Problem}
\label{sec:ASP-formulation}

%In this section, the ASP formulations that are developed will be explained.
%For an FA $A=(Q,\Sigma,\delta)$ and a constant $c$,
%an ASP formula is constructed that is satisfiable if and only if $A$
%has a synchronizing word $w$ of length $c$. Also, we present ASP
%formulations where a range is given for the value of $c$, and
%the ASP solver attempts to minimize $c$ such that $A$ has a
%synchronizing sequence of length $c$.
%In all ASP formulations, we use the following atoms.

Let us first consider the decision version of the shortest
synchronizing sequence problem: For an FA $A=(Q,\Sigma,\delta)$ and
a positive integer constant $c$, decide whether $A$ has a synchronizing word $w$ of
length $c$.

Without loss of generality, we represent states and input symbols of
an FA $A=(Q,\Sigma,\delta)$, by the range of numbers $1..n$ and
$1..k$ ($n=|Q|$, $k=|\Sigma|$), respectively. Then an FA
$A=(Q,\Sigma,\delta)$ can be described in ASP by three forms of atoms 
given below:
\begin{itemize}
\item
$\ii{state}(s)$ ($1\leq s \leq n$) describing the states in $Q$,
\item
$\ii{symbol}(j)$ ($1\leq j \leq k$) describing the input symbols in
$\Sigma$, and
\item
$\ii{transition}(s,j,s')$ ($1\leq s,s' \leq n$, $1\leq j \leq k$)
describing the transitions $\delta(s,j)=s'$.
\end{itemize}
We represent possible lengths $i$ of sequences by atoms of the form
$\ii{step}(i)$ ($1 \leq i \leq c$).

A synchronizing sequence of length $c$ is characterized by atoms of
the form $\ii{synchro}(i,x)$ ($1\leq i \leq c$, $1 \leq x \leq k$)
describing that the $i$'th symbol of the word is $x$.

Using these atoms, we can represent the decision version of the
shortest synchronizing sequence problem with a ``generate-and-test''
methodology used in various ASP formulations. In the following, we
present two different ASP formulations based on this approach.

In these ASP formulations, we use an auxiliary concept of a {\em
path in $A$ characterized by a sequence $w_1,w_2,\dots,w_x$ of
symbols in $\Sigma$}, which is defined as a sequence $q_1, q_2,
\dots, q_{x+1}$ of states in $Q$ such that $\delta(q_i,w_i)=q_{i+1}$
for every $i$ ($1\leq i \leq x$). The existence of such a path of
length $i$ in $A$ from a state $s$ to a state $q$ (i.e., the
reachability of a state $q$ from a state $s$ by a path of length $i$
in $A$) characterized by the first $i$ symbols of a word $w$ is
represented by atoms of the form $\ii{path}(s,i+1,q)$ defined as
follows:

\beq
\ba l
\ii{path}(s,1,s) \lar \ii{state}(s) \\
\ii{path}(s,i+1,q) \lar \ii{path}(s,i,r), \ii{synchro}(i,x), \ii{transition}(r,x,q), \\
\qquad \ii{state}(s), \ii{state}(r), \ii{state}(q), \ii{symbol}(x),
\ii{step}(i)
\ea
\eeq{eq:path}

%Intuitively, $synchro(I, X)$ means that $I^{\mbox{th}}$ element of
%the synchronizing sequence is the input symbol $X$. This atom is
%used to generate/guess an input sequence. The constraints that will
%be given below forces this sequence to be a synchronizing sequence.
%As $synchro$ guesses an input sequence, we keep track of the state
%of $A$ that is reached at each step and from each initial state when
%the guessed input sequence is applied. This is performed by using
%another atom called $path$. Intuitively, $path(Qi, I, Qf)$ indicates
%that, if the initial state is $Qi$, after applying the first $I$
%input symbols guessed by $synchro$, we reach to state $Qf$.
%
%The constraints applied on the atoms $synchro$ and $path$ will be explained
%below.

\subsection{Connecting All States to a Sink State}
\label{sec:ASP-collectAll}

In the first ASP formulation, which we call \aspI, first we
``generate'' a sequence $w$ of $c$ symbols by the following choice
rule:

\beq
1\{\ii{synchro}(i,j): \ii{symbol}(j)\}1 \lar \ii{step}(i)
\eeq{eq:generate}

\noindent where $\ii{step}(i)$ is defined by a set of facts:

\beq
\ii{step}(i) \lar \qquad (1\leq i \leq c)
\eeq{eq:step1}

Next, we ensure that it is a synchronizing sequence by ``testing''
that it does not violate the condition:

\begin{itemize}
\item[$C_1$] There exists a sink state $f\in Q$ such that
every path in $A$ characterized by $w$ ends at $f$.
\end{itemize}

\noindent by adding the following constraints

\beq
\ba l
\lar \ii{sink}(f), \no\ \ii{path}(s,c+1,f),
\ii{state}(s), \ii{state}(f)
\ea
\eeq{eq:test}

\noindent where $\ii{sink}(f)$ describes a sink state:

\beq
1 \{\ii{sink}(f) : state(f)\} 1 \lar .
\eeq{eq:sink}

The union of the program \aspI\ that consists of the
rules~(\ref{eq:generate}), (\ref{eq:step1}), (\ref{eq:test}),
(\ref{eq:sink}), (\ref{eq:path}), with a set of facts describing an
FA $A$ has an answer set iff there exists a synchronizing sequence
of length $c$ for $A$.

\subsection{Merging States Pairwise}
\label{sec:ASP-pairwise}

In the second ASP formulation, which we call \aspII, first we
``generate'' a sequence $w$ of $c$ symbols by the choice
rule~(\ref{eq:generate}).

Next, we ensure that it is a synchronizing sequence by ``testing''
that it does not violate the following condition, instead of
constraint $C_1$:

\begin{itemize}
\item[$D_1$] For every pair $q_i,q_{i+1}$ of states in
$Q= \{q_1, q_2, \ldots, q_n\}$, $\delta(q_i,w) = \delta(q_{i+1},w)$.
\end{itemize}

\noindent by adding the following cardinality constraints

\beq \lar 1 \{\ii{merged}(r): \ii{state}(r), r < n\} n-2
\eeq{eq:test2}

\noindent where $\ii{merged}(r)$ describes that there exists a state
$s$ reachable from the states $r$ and $r+1$ by paths characterized
by the first $i$ symbols of $w$ for some $i$ ($1 \leq i \leq c$):

\beq
\ba l
\ii{merged}(r) \lar \ii{path}(r,i,s), \ii{path}(r+1,i,s), \\
\qquad \ii{state}(s), \ii{state}(r), \ii{state}(r+1), \ii{step}(i)
\ea
\eeq{eq:merged}

The union of the program \aspII\ that consists of the
rules~(\ref{eq:generate}), (\ref{eq:step1}), (\ref{eq:test2}),
(\ref{eq:merged}), (\ref{eq:path}), with a set of facts describing
an FA $A$ has an answer set iff there exists a synchronizing
sequence of length $c$ for $A$.

\subsection{Optimization}
\label{sec:ASP-optimization}

The ASP formulations given in Section~\ref{sec:ASP-collectAll} and
Section~\ref{sec:ASP-pairwise}, with a set of facts describing an FA
$A$, have answer sets if the given FA $A$ has a synchronizing
sequence of length $c$. In order to find the length of the shortest
synchronizing sequence, one can perform a binary search on possible
values of $c$.

In this section, we present another ASP formulation where we let the
ASP solver first decide the length $l$ of a shortest synchronizing
sequence, where $l \leq c$:

\beq
1\{\ii{shortest}(l): 1 \leq l \leq c\} 1 \lar
\eeq{eq:shortest}

\noindent and declare possible lengths of sequences:
\beq
\ii{step}(j) \lar \ii{shortest}(i) \qquad (1 \leq j \leq i \leq c) .
\eeq{eq:step2}

Next, we ensure that $l$ is indeed the optimal value, by the
following optimization statement

\beq
\# \ii{minimize} [ \ii{shortest}(l) = l ] \lar
\eeq{eq:minimize}

We denote by \aspIopt\ (resp. \aspIIopt) the ASP formulation
obtained from \aspI\ (resp. \aspII) by adding the rules
(\ref{eq:shortest}) and (\ref{eq:minimize}), and replacing the rules
(\ref{eq:step1}) by the rules (\ref{eq:step2}). If \aspIopt\ (resp.
\aspIIopt) with a set of facts describing an FA $A$ has an answer
set $X$ then $X$ characterizes a shortest synchronizing sequence for
$A$.

% % % % % % % % % % % % % % % % % % % % % % % % % % %

\section{SAT Formulation of the Shortest Synchronizing Sequence
Problem}
\label{sec:SAT}

In~\cite{Skv}, a SAT formulation of the problem of checking if an FA
$A$ has a synchronizing sequence of a certain length is presented.
However, this formulation is given only for FA with two input
symbols. We extend this SAT formulation to FA with any number of input
symbols as follows.

We first define a boolean operator $\nabla$ that will simplify the
description of our SAT formulation. For a given set of boolean
variables $\{ r_1, r_2, \ldots, r_k \}$, we define $\nabla \{r_1 ,
r_2 , \cdots , r_k \}$ as follows:

\begin{center}
   $\nabla$ \{$r_1$,$r_2$,$\cdots$,$r_k$\} $\equiv$  (($r_1$ $\Rightarrow$ ( $\neg$ $r_2$ $\wedge$  $\neg$ $r_3$ $\wedge$  $\cdots$     $\wedge$  $\neg$ $r_k$))$\wedge$
($r_2$ $\Rightarrow$ ( $\neg$ $r_1$ $\wedge$  $\neg$ $r_3$ $\wedge$
$\cdots$     $\wedge$  $\neg$ $r_k$))$\wedge$ $\cdots$ ($r_k$
$\Rightarrow$ ( $\neg$ $r_1$ $\wedge$  $\neg$ $r_2$ $\wedge$ $\cdots$
$\wedge$  $\neg$ $r_{k-1}$))$\wedge$ ($r_1$ $\vee$ $r_2$ $\vee$
$\cdots$  $\vee$ $r_k$ ))
\end{center}

\noindent Intuitively, $\nabla \{r_1 , r_2 , \cdots , r_k \}$ is
true with respect to an interpretation $I$ iff exactly one of the
variables $r_i$ is true and all the others are false with respect to
$I$.

Checking the existence of a synchronizing sequence of length $c$ is
converted into a SAT problem by considering the following boolean
formulae. Below we use the notation $[c]$ to denote the set $\{1,
2, \ldots, c\}$.

   \begin{itemize}
   \item $F_1$: An input sequence of length $c$ has to be created.
   At each step of this input sequence, there should be exactly one input symbol being used.
   For this purpose, we generate Boolean variables $X_{l,x}$ which should be
   set to true (by an interpretation) only if at step $l$ the input symbol $x$ is used.
   The following formulae make sure that only one input symbol is picked for each step $l$.\\

   \begin{quote}
     $\sigma_1  =  \bigwedge_{l\in[c]} ( \nabla \{ X_{l,x}  | x \in X \} )$\\
   \end{quote}

   \item $F_2$: Similar to what we accomplish in ASP formulations by atoms of the form
   $\ii{path}(s,i,q)$,
   we need to trace the state reached when the input sequence guessed by formula $\sigma_1$ is applied.
   For this purpose, boolean variables $S_{i,j,k}$ (which we call {\em state tracing variables})
   are created which are set to true (by an interpretation)
   only if we are at state $q_k$ at step $j$ when we start from state $q_i$. We first
   make sure that for each starting state and at any step, there will always be exactly one
   current state.\\

    \begin{quote}
       $\sigma_2 = \bigwedge_{i \in [n], l \in [c]} ( \nabla \{ S_{i,l,j}  | j \in n \} )$\\
    \end{quote}

   \item $F_3$: The initial configuration of the FA $A$ must be
   indicated. For this purpose state tracing variables should be
   initialized for their first step.\\

\begin{quote}
  $\sigma_3 = \bigwedge_{i \in [n]} ( S_{i,1,i} )$\\
\end{quote}

   \item $F_4$: Again, similar to the constraints in ASP formulations,
   over atoms of the form $\ii{path}(s,i,q)$, we have the corresponding SAT formulae
   to make sure that state tracing variables are assigned according to the transitions of the FA $A$.
   For each state $q_j$ and input $x$ of $A$, if we have $\delta(q_j,x)=q_k$ in $A$, then we
   generate the following formulae:\\

\begin{quote}
  $\sigma_4 = \bigwedge_{i,j\in[n], l\in[c], x \in X } (( S_{i,l,j} \bigwedge
            X_{l,x}) \Rightarrow S_{i,l+1,k} )$\\
\end{quote}

   \item $F_5$: A synchronizing sequence $w$ merges all the states at a sink state
   after the application of $w$. We use boolean variable $Y_{i}$ to pick a sink state. Since only one of the
   states has to be a sink state, we introduce the following formulae:\\

   \begin{quote}
            $\sigma_5 = ( \nabla \{ Y_{i} | i \in [n]  \} )$\\
   \end{quote}

   \item $F_6$: Finally, we need to make sure that all the states reach the sink state
   picked by $F_5$ at the end of the last step after the application of the synchronizing sequence
   guessed by formulae $F_1$.\\

\begin{quote}
  $\sigma_6 = \bigwedge_{i,j \in [n]} ( Y_{i} \Rightarrow  S_{j,c+1,i})$\\
\end{quote}

The conjunction of all formulae introduced above is a Boolean
formula that is satisfiable iff there exists a synchronizing
sequence of FA $A$ of length $c$.

\end{itemize}

% % % % % % % % % % % % % % % % % % % % % % % % % % %

\section{Experimental Study}
\label{sec:experiments}

In this section, we present the experimental study carried out to compare
the performance of the ASP formulations, the SAT formulation, and the brute--force
algorithm for generating a shortest synchronizing sequence.

We first present our experiments using finite automata that are generated randomly.
Given the number of states and the number of input symbols,
an FA is generated by assigning the next state of each transition
randomly. If the FA generated in this way does not have a synchronizing
sequence, then it is discarded. Otherwise, it is included in the
set of FAs used in our experiments. We generated 100 random FAs this way
for each number of states we used in the experiments (except for the
biggest set of tests with 50 states and 4-6 input symbols, where we
use only 50 FAs to speed up the experiments).

The implementation of the brute--force algorithm in the tool
COMPAS~\cite{Chmiel:2010:CCP:1964285.1964294} is used. The
brute--force algorithm could be used for FAs with up to 27 states.
Beyond this number of states, COMPAS could not complete the
computation due to memory restrictions.

We implemented tools that create ASP and SAT formulations from
a given FA and an integer constant $c$ as explained in
Section~\ref{sec:ASP-formulation}, Section~\ref{sec:SAT}, and
also the SAT formulation given in~\cite{Skv} for FAs with
two inputs only.

In the results given below, the formulations \aspI, \aspII,
\aspIopt, and \aspIIopt~refer to the ASP formulations
given in Section~\ref{sec:ASP-formulation}.
\satI~and \satII~refer to the SAT formulations given in~\cite{Skv} and
Section~\ref{sec:SAT}, respectively. \bforce~refers to the brute--force algorithm.

Note that the ASP formulations \aspIopt~and \aspIIopt~report the
length of a shortest synchronizing sequence, provided that the
constant $c$ given is not smaller than the length of a shortest
synchronizing sequence. When $c$ is not big enough, another
experiment is performed by doubling the value of $c$. We report
only the results from successful \aspIopt~and \aspIIopt~experiments,
where a sufficiently large $c$ is given.
An experimental study is presented in \cite{Skv} where the length of the 
shortest synchronizing sequence is reported to be around $2 \sqrt{n}$ on
the average for an $n$ state automaton with two input symbols. We 
therefore initially take the value of $c$ as $2 \sqrt{n}$. 

On the other hand, the ASP formulations \aspI~and \aspII, and also
the SAT formulations \satI~and \satII, only report if a
synchronizing sequence of length $c$ exists or not. Therefore, one
has to try several $c$ values to find the length of the shortest
synchronizing sequence using these formulations. In our experiments
with these formulations, we find the length of a shortest
synchronizing sequence by applying a binary search on the value of
$c$, by using a script that invokes the ASP solver for each attempt
on a possible value of $c$ separately. We similarly take the initial
value of $c$ to be $2 \sqrt{n}$ as explained above. The time reported 
is the total time taken by all the attempts
until the length of a shortest synchronizing sequence is found. The
memory reported is the average memory usage in these attempts.

The experiments are carried out using MiniSat 2.2.0~\cite{eenS03}
and Clingo 3.0.3~\cite{gekakaosscsc11a} running on Ubuntu Linux
where the hardware is a 2.4Ghz Intel Core-i3 machine.

In Table~\ref{table1} and Table~\ref{table2}, the time and the memory
performance of the formulations \aspI, \aspII, \satI, and the
brute--force algorithm are given.
We could not get a report on the memory usage of COMPAS for the brute--force
algorithm, hence no data is given for the brute--force algorithm in
Table~\ref{table2}. In this set of experiments, the number of states $n$
ranges between 5 and 27, and the number of input symbols is fixed to 2.

\begin{table}[t]
%\begin{center}
\centering
\caption{Experiments on FAs with 2 input symbols (time - secs)}
\label{table1}

\begin{tabular}{r@{\qquad}r@{\qquad}r@{\qquad}r@{\qquad}r@{\qquad}r@{\qquad}r}
\hline\noalign{\smallskip} $n$ & \aspI & \aspII   & \aspIopt & \aspIIopt & \satI&  \bforce \\
\noalign{\smallskip} \hline \noalign{\smallskip}
         5 & 0 &  0 & 0 &  0 &  7 &   0 \\
        10 & 2 &  2 & 2 & 11 & 10 &   0 \\
        15 & 1 &  2 & 2 &  5 & 10 &   0 \\
        20 & 4 &  6 & 5 &  8 & 12 &   4 \\
        25 & 6 & 12 & 7 & 14 & 13 &  73 \\
        26 & 7 & 11 & 8 & 15 & 14 & 145 \\
        27 & 9 & 12 & 8 & 14 & 15 & 289 \\
\hline
\end{tabular}
%\end{center}
\end{table}

\begin{table}[t]
\centering
%\begin{center}
\caption{Experiments on FAs with 2 input symbols (memory - kBytes)}
\label{table2}
\begin{tabular}{r@{\qquad}r@{\qquad}r@{\qquad}r@{\qquad}r@{\qquad}r}
\hline\noalign{\smallskip} $n$ & \aspI & \aspII & \aspIopt & \aspIIopt & \satI\\
\noalign{\smallskip} \hline \noalign{\smallskip}
        5  & 7750 & 7731 & 7622     & 7620 & 7677 \\
        10 & 8169 & 7278 & 8154     & 8160 & 7983 \\
        15 & 6284 & 6566 & 6465     & 6300 & 7847 \\
        20 & 6909 & 6911 & 6943     & 6947 & 7810 \\
        25 & 6769 & 6775 & 6937     & 6779 & 8066 \\
        26 & 7151 & 6798 & 7136     & 6822 & 8213 \\
        27 & 7123 & 7106 & 6744     & 7119 & 8113 \\
\hline
\end{tabular}
%\end{center}
\end{table}

In Table~\ref{table3} and Table~\ref{table4}, the time and the memory
performance of the formulations \aspIopt, \aspIIopt, and \satII~are
given on FAs with the number of states $n \in \{30, 40, 50\}$ and the number of
inputs $k \in \{2, 4, 6\}$.

\begin{table}[t]
\centering
%\begin{center}
\caption{Experiments on FAs with different number of inputs (time - secs)}
\label{table3}
\begin{tabular}{r@{\qquad}r@{\qquad}r@{\qquad}r@{\qquad}r@{\qquad}r@{\qquad}r}
\hline\noalign{\smallskip} $n$ & $k$ & \aspI & \aspII & \aspIopt & \aspIIopt  & \satII \\
\noalign{\smallskip} \hline \noalign{\smallskip}
        30 & 2 &    4 &   17 &    4 &   18 &   49 \\
        30 & 4 &   66 &   80 &   45 &   57 &  101 \\
        30 & 6 &  208 &  405 &  160 &  231 &  490 \\
        40 & 2 &   33 &   45 &   71 &  122 &  222 \\
        40 & 4 &  348 &  380 &  244 &  311 &  472 \\
        40 & 6 & 1158 & 1400 &  707 &  980 & 2133 \\
        50 & 2 &   93 &  120 &  117 &  146 &  430 \\
        50 & 4 &  902 & 1101 &  835 &  833 & 2975 \\
        50 & 6 & 3205 & 4010 & 2705 & 3032 & 7492 \\
\hline
\end{tabular}
%\end{center}
\end{table}

\begin{table}[t]
\centering
%\begin{center}
\caption{Experiments on FAs with different number of inputs (memory - kBytes)}
\label{table4}
\begin{tabular}{r@{\qquad}r@{\qquad}r@{\qquad}r@{\qquad}r@{\qquad}r@{\qquad}r}
\hline\noalign{\smallskip} $n$ & $k$ & \aspI & \aspII & \aspIopt & \aspIIopt  & \satII \\
\noalign{\smallskip} \hline \noalign{\smallskip}
        30 & 2 & 6063 & 7143 &  4973 &  6140 & 42764 \\
        30 & 4 & 7309 & 7438 &  5278 &  7457 & 49681 \\
        30 & 6 & 7735 & 7621 &  7496 & 10457 & 52250 \\
        40 & 2 & 6709 & 6029 &  5616 &  7448 & 67362 \\
        40 & 4 & 7050 & 7073 &  7880 &  8550 & 78697 \\
        40 & 6 & 7764 & 8024 &  7983 & 10234 & 84671 \\
        50 & 2 & 7222 & 8336 &  8965 & 16072 & 86453 \\
        50 & 4 & 8438 & 9056 &  9931 & 12843 & 85157 \\
        50 & 6 & 8773 & 9228 & 10729 & 14092 & 93118 \\
\hline
\end{tabular}
%\end{center}
\end{table}

Table~\ref{table1} shows that the brute--force approach uses much
more time than the other approaches, especially as the size of the FA
gets bigger. Therefore, after a certain threshold size, the brute--force
approach is not an alternative.

By investigating the results given in Table~\ref{table1} and
Table~\ref{table3}, one can see that the ASP formulation approach of 
\aspI~and \aspIopt~perform better than \aspII~and \aspIIopt, in general.
This may be due to that the number of ground instances of (4) and (5)
is smaller than that of (6) and (7). However, after intelligent
grounding of Clingo, the program sizes of \aspIopt~and \aspIIopt~become
comparable. On the other hand, we have observed that \aspIIopt~leads to
more backtracking and restarts compared to~\aspIopt. For example, for
an instance of 50 states and 2 input symbols, \aspIopt~leads to 82878
choices and no restarts, whereas \aspIIopt~leads to  137276 choices and
4 restarts. This may be due to that, in \aspIopt with respect to (4)
and (5) , once a sink node is selected, for every state, existence of
a path of a fixed length is checked; on the other hand, in \aspIIopt
with respect to (6) and (7), for every state, existence of two paths
of the same length is checked, which may intuitively lead to more
backtracking and restarts.
On the other hand, the memory performances of all ASP approaches are similar,
as displayed by Table~\ref{table2} and Table~\ref{table4}.

As for the comparison of the ASP and SAT approaches, one can see that
the ASP approaches are both faster and uses less memory than the SAT
approach, in general. However, the ASP approach seems to have a
faster increase in the running time compared to the SAT approach.
This trend needs to be confirmed by further experiments.

We also experimented with finite state automata from MCNC'91 benchmarks~\cite{Yang91}.
We used only those finite state machine examples in this benchmark set that 
correspond to completely specified finite state automata. The results of 
these experiments are given in Table~\ref{table5} for the time comparison.
We obtained similar results to what we have observed in our experiments
on random finite state automata. The time performance of the ASP approaches 
are better than the SAT approach in these experiments as well.
We note that the benchmark example ``dk16'', which is also the automaton
having the largest number of states, has the 
longest running time among the automata in the benchmark set. 
However, the running time does not depend only on the number of
states. The number of input symbols and the length of the 
shortest synchronizing sequence would also have an effect.
For the comparison of the memory used for these examples, 
all ASP approaches used around 6 MBytes of memory, whereas the 
SAT approach used minimum 6 MBytes and maximum 38 MBytes memory
for these experiments.

\begin{table}[t]
\centering
%\begin{center}
\caption{Experiments on FAs from MCNC benchmarks (time - msecs)}
\label{table5}
\begin{tabular}{l@{\qquad}r@{\qquad}r@{\qquad}r@{\qquad}r@{\qquad}r@{\qquad}r@{\qquad}r}
\hline\noalign{\smallskip} Name & $n$ & $k$ & \aspI & \aspII & \aspIopt & \aspIIopt  & \satII \\
\noalign{\smallskip} \hline \noalign{\smallskip}
bbtas    &  6 & 4 &   15 &   18 &   10 &   14 &    52 \\
beecount &  7 & 8 &   18 &   18 &   10 &    9 &    89 \\
dk14     &  7 & 8 &   18 &   20 &   15 &   17 &    62 \\
dk15     &  4 & 8 &   13 &   14 &    7 &    7 &    20 \\
dk17     &  8 & 4 &   27 &   26 &    8 &    8 &    69 \\
dk27     &  7 & 2 &   22 &   21 &    6 &    6 &    45 \\
dk512    & 15 & 2 &   33 &   27 &   11 &   12 &   278 \\
dk16     & 27 & 4 &  191 &  231 &  132 &  127 & 21253 \\
lion9    &  9 & 4 &   91 &  138 &   36 &  137 &   449 \\
MC       &  4 & 8 &   18 &   18 &    7 &    7 &    53 \\
\hline
\end{tabular}
%\end{center}
\end{table}

% % % % % % % % % % % % % % % % % % % % % % % % % % %

\section{Conclusion and Future Work}
\label{sec:conclusion}

In this paper, the problem of finding a shortest synchronizing
sequence for a FA is formulated in ASP. Four different ASP formulations 
are given. Also an extension of the SAT
formulation of the same problem given in~\cite{Skv} is suggested.

The performance of these formulations are compared by an
experimental evaluation. The ASP and SAT formulations are shown to
scale better than the brute--force approach. The experiments
indicate that the ASP formulations perform better than the SAT
approach. However this needs to be further investigated with an
extended set of experiments.

Based on the encouraging results obtained from this work, using ASP
to compute some other special sequences used in finite state machine
based testing can be considered as a future research direction. For
example checking the existence of, and computing a Preset
Distinguishing Sequence is a PSPACE--hard problem~\cite{LY94}.
Although checking the existence and computing an Adaptive
Distinguishing Sequence~\cite{LY94} can be performed in polynomial
time, generating a minimal Adaptive Distinguishing Sequence is an
NP--hard problem. These hard problems can be addressed by using ASP.

% % % % % % % % % % % % % % % % % % % % % % % % % % %

\end{document}